\documentclass[letterpaper]{article}
\usepackage{aaai2026}
\usepackage{times}
\usepackage{helvet}
\usepackage{courier}
\usepackage[hyphens]{url}

\usepackage{graphicx}
\usepackage{natbib}
\usepackage{caption}
\usepackage{amsmath}
\usepackage{amssymb}
\usepackage{booktabs}
\definecolor{paperlinkblue}{HTML}{1F4E79}
\newcommand{\paperlink}[2]{\leavevmode\pdfstartlink attr{/Border [0 0 0]} user{/Subtype /Link /A << /S /URI /URI (#1) >>}\textcolor{paperlinkblue}{#2}\pdfendlink}
\copyrightyear{2026}
\AddToHookNext{shipout/foreground}{%
  \put(0,-42.27pt){\makebox[\paperwidth]{\normalfont\fontsize{8pt}{10pt}\selectfont The Fortieth AAAI Conference on Artificial Intelligence (AAAI-26)}}%
}

\title{Learning Object-Centric Motion Priors from Human for Robotic Dexterous Manipulation}
\author{
Zhengdong Hong, Guofeng Zhang\thanks{Corresponding author.}
}
\affiliations{
State Key Lab of CAD\&CG, Zhejiang University\\
    12321092@zju.edu.cn, zhangguofeng@zju.edu.cn
}

\begin{document}

\maketitle

\begin{abstract}

Manipulating diverse objects with multi-fingered dexterous hands is challenging due to the high dimensionality and complex dynamics. Human-Object Interaction (HOI) datasets provide rich knowledge about task information and embodied interactions. Instead of solely imitating the human demonstrations, our method learns to holistically predict future hand-object states by leveraging these datasets. The predicted future states of the object can serve as a general-purpose reward term for reinforcement learning, reducing reliance on task-specific reward engineering and enhancing generalization across tasks. We conduct extensive experiments on three manipulation tasks in simulation and the real world. Our approach outperforms existing SOTA methods in both success rate and generalizability on novel objects. Furthermore, we validate the cross-embodiment compatibility of our methods by successfully deploying the skills on different robot hands.\end{abstract}

{\small
\noindent\textit{Accepted as an oral paper at AAAI 2026.}\\
\textbf{Website:} \paperlink{https://ojs.aaai.org/index.php/AAAI/article/view/38892}{AAAI proceedings}\\
\textbf{Video:} \paperlink{https://ojs.aaai.org/index.php/AAAI/article/view/38892/47664}{Oral presentation}\par
}

\section{Introduction}
Achieving human-level dexterous manipulation is a critical but challenging goal for robotics, primarily due to the high degrees of freedom in robot hands and complex contact dynamics. Previous methods, such as model-based trajectory optimization \cite{pang2023global, jin2024complementarity}, struggle with these dynamic complexities. Meanwhile, model-free reinforcement learning (RL) \cite{rajeswaran2017learning, gupta2016learning, akkaya2019solving, chen2023visual, wang2024lessons} is often inefficient, demanding extensive exploration and task-specific reward engineering. While imitation learning from robot data \cite{chi2023diffusion} presents a viable solution, the required large-scale data collection is frequently impractical for research labs \cite{wang2024dexcap, black2410pi0}. In contrast, the vast amount of available human data, which implicitly encodes rich manipulation experience, offers a promising and scalable alternative for learning dexterous manipulation.

What should we learn from the human demonstration? To answer this, let's look at how humans learn. Children learn to interact with the physical world using the following two-stage learning paradigm. The first stage involves forming an abstract task concept through watching others play. In the second stage, this conceptual knowledge is grounded and refined through physical experimentation by themselves, facilitating the development of operational competence. Similarly, prior literature~\cite{qin2022dexmv,chen2024vividex,shi2025learning, chen2024object, li2025maniptrans} investigates a similar two-stage pipeline for learning dexterous manipulation skills. Most of them either focus solely on imitating human motions or designing task-specific rewards, then testing on only one task, which makes it hard to draw a general conclusion from that. Some of these works overlook an important fact: while humans are manipulating objects, they are also subconsciously predicting the future state changes of the objects. This ability to predict the future state of objects not only helps humans understand tasks but also enables them to perform more general operations on different tasks. Inspired by that, our observation is that considering human-object as a whole and learning human manipulation priors from hand-object interactions provides more general and efficient task-level guidance for the second-stage low-level policy learning. Instead of using human-designed task-specific rewards like lifting rewards in previous work~\cite{chen2024vividex, shi2025learning}, we believe that human-object interaction priors encourage robots to manipulate the object in the way described in the dataset, which is task-agnostic to different tasks.

Specifically, we propose a pipeline to auto-regressively predict future hand-object states holistically. And we use the predicted hand and object states to guide the low-level policy learning in different tasks without task-specific reward design in low-level reinforcement learning in a physical simulation. Then we perform zero-shot sim-to-real skill transfer to the real-world robots. Our contributions are as follows:

\begin{itemize}
    \item We propose an efficient learning framework to utilize human-object interaction priors for robot skill learning, without the need for real-robot data.
    \item Predicting hand-object priors holistically and properly utilizing the predicted object states can not only facilitate reinforcement learning by alleviating the heavy reliance on task-specific reward design, but also enable our pipeline to be more general to various task settings.
    \item We conducted extensive real-robot experiments on various tasks, embodiments, and object settings, outperforming existing SOTA methods in all three challenging tasks.

\end{itemize}

\begin{figure*}[htb]
\centering
\includegraphics[width=\textwidth]{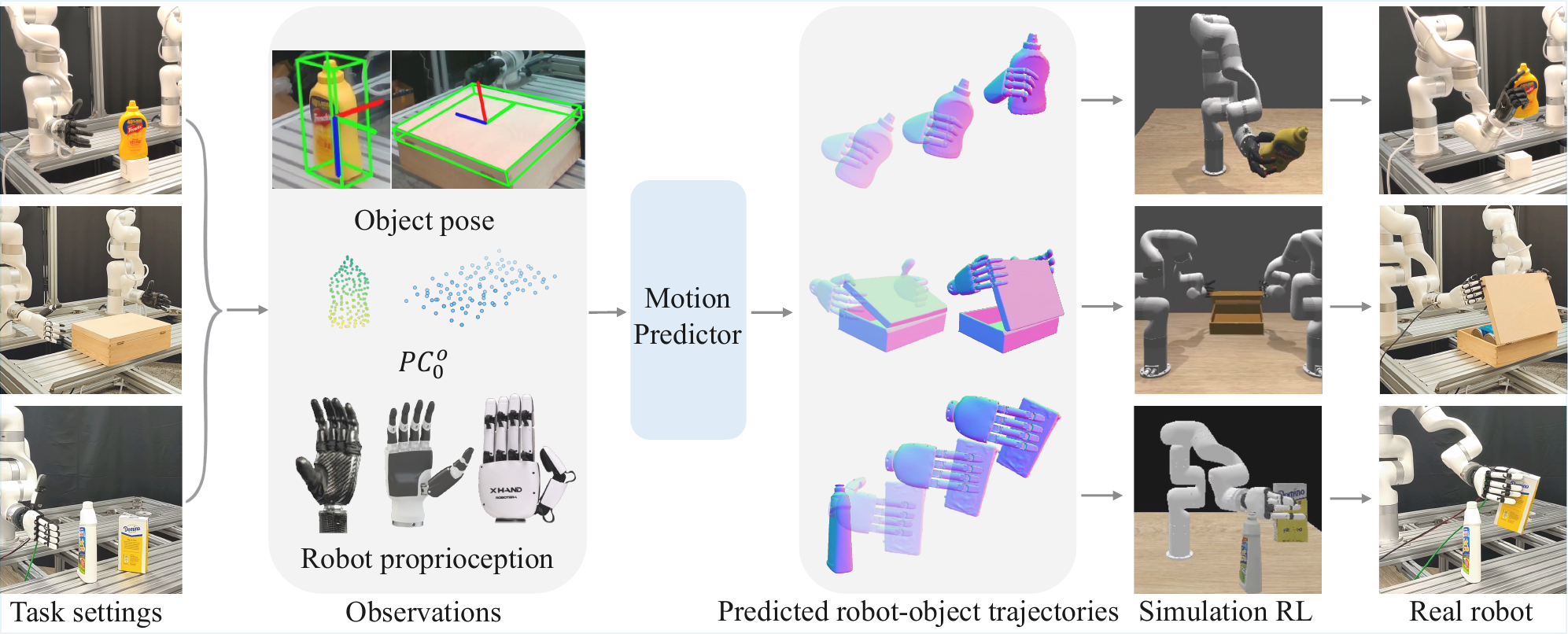}
\caption{Our method begins by extracting observations from the initial robot-object configuration. These observations are input into a motion predictor to generate a reference trajectory. Guided by this trajectory, we employ reinforcement learning to train the target policy. Finally, the learned policy is deployed directly to the physical robot via zero-shot Sim-to-Real transfer. Notably, since the learned priors are general to hand morphologies, our framework is adaptable to various dexterous hands.}
\label{fig: pipeline}
\end{figure*}

\begin{figure*}[htb]
\centering
\includegraphics[width=\textwidth]{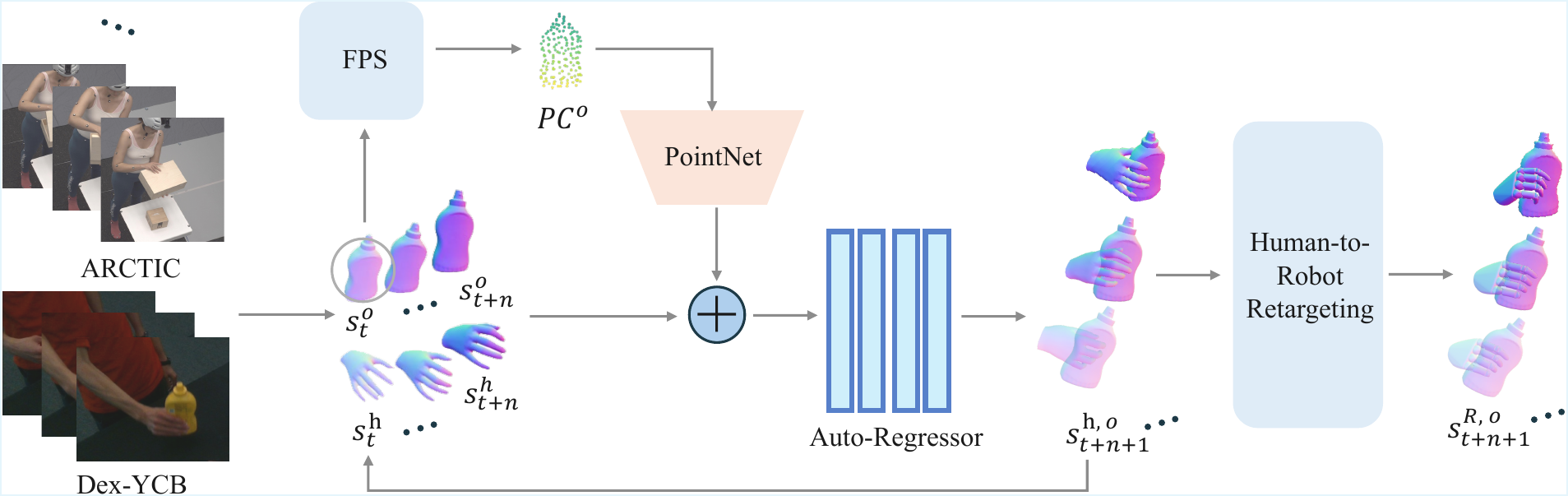}
\caption{Illustration of components in our HOI motion predictor.}
\label{fig: regressor}
\end{figure*}

\section{Related Work}

\subsection{Dexterous Manipulation}
Dexterous manipulation is complicated by the high dimensionality of the end effector. This complexity poses a problem for traditional trajectory optimization techniques \cite{rus1999hand, mordatch2012contact, wu2022learning}, as they require precise dynamic models that are impractical for real-world applications.
Learning-based methods address this by learning an observation-to-action policy directly from data, but they introduce new challenges. Reinforcement learning approaches \cite{qi2023general, yuan2024robot, wang2024lessons} can be sample-inefficient and depend on meticulous reward engineering. On the other hand, imitation learning methods \cite{bain1995framework, chi2023diffusion}, which use data from teleoperation \cite{qin2023anyteleop, wang2024dexcap}, are often constrained by the expense of data collection and the quality of the expert demonstrations. Vision-Language-Action methods~\cite{pan2024vision, zhong2025dexgraspvla} aim at utilizing pretrained language models to boost the network performance, while only demonstrating tasks not harder than pick-and-place. Grasp synthesis methods~\cite{ye2025dex1b,wan2023unidexgrasp++} are mainly tested in simulation, and the potential in more challenging tasks beyond pick-and-place is not fully investigated. Our methods combine the merits of using human data and scaled-up simulation, enabling real-robot dexterous manipulation with high success rates.

\subsection{Learning Dexterous Manipulation from Human}
Internet human video data contains abundant information about human experiences. Previous research \cite{mendonca2023structured, ma2022vip, nair2022r3m} seeks to learn general visual representations from internet-scale human video data to guide robot learning. However, those methods only showcase results on a 1-DoF robot gripper since the features trained on 2D human videos lack the 3D information necessary for learning high-dimensional skills like dexterous manipulation. Different from internet video human data, human-object interaction (HOI) data \cite{chao2021dexycb, fan2023arctic, liu2022hoi4d} are usually collected by multi-view RGBD cameras, with extracted and labeled 3D human pose estimations and 6-DoF object pose estimations, which contain abundant 3D information necessary for learning dexterous manipulation. Given the 3D motions of the human hand, an intuitive idea is to directly transfer human motions to robot actions. \cite{qin2023anyteleop} proposed a retargeting system based on inverse kinematics to map the human hand motions to robot hand states and actions. \cite{qin2022dexmv,liu2025dextrack} further prove that the retargeted trajectories can be used to train imitation learning policies in different tasks.

Following them, \cite{chen2024vividex, zhou2024learning, shi2025learning} apply trajectory augmentation and train a state-based policy based on that. Different from direct mapping, \cite{chen2024object} proposes to train a human motion prediction network given a target object trajectory, then augment and refine it in physical simulation using Reinforcement Learning (RL). However, defining target object trajectories in open world requires extra efforts. \cite{li2025maniptrans} train a 2-stage process imitating human, but their input is fixed object initial poses with object pose randomization as 0, constraining open-world applications.

To address this, \cite{zhao2024dexh2r} introduces human-in-the-loop adjustments in generating reference trajectories, but it requires human labor.
\cite{singh2024hand} requires real-world robot demonstrations or RL to finetune the policy. Most of them focus on a single task and can hardly generalize to novel object configurations. We demonstrate our applicability in various challenging tasks under different object configurations.

\section{Methods}
Our proposed pipeline is illustrated in Fig.~\ref{fig: pipeline}. The process begins by extracting observations from the initial state of a real-world robot and object scene. These observations are then fed into a motion predictor to generate a reference trajectory. Using this trajectory as guidance, we employ a reinforcement learning framework to learn the desired skill policy. Finally, this learned policy is transferred directly to the physical robot using a zero-shot sim-to-real approach.
\subsection{Hand-Object Poses Prediction}
Fig.~\ref{fig: regressor} is our hand-object-interaction (HOI) motion predictor. We use Dex-YCB~\cite{chao2021dexycb} and ARCTIC~\cite{fan2023arctic} as our datasets. Both of them have well-labeled 6DoF hand-object pose pairs for each timestamp, noted as $\hat{s_{t}^h}$ which is represented by MANO~\cite{romero2022embodied} hand poses and $\hat{s_{t}^o}$ which are the quaternion $\mathbf{q_{t}^o}$ and the translation $\mathbf{T_{t}^o}$ of a object pose. We use a coordinate system with the geometric center of the object at $t=0$ as the origin. For articulated objects like a laptop and a box, we record the relative pose of the upper part to the lower part (base). For each motion sequence, we first use farthest point sampling (FPS)~\cite{qi2017pointnet++} to extract a 3D point cloud $PC^o$ consisting of 100 points from the object mesh, which is later processed by a PointNet~\cite{qi2017pointnet} to get a feature about the object shape information. We use a 6-layer GPT-2 transformer~\cite{radford2019language} to predict future hand-object pose pairs $s_{t+n+1}^{h, o}$ using the history observations $s_{t}^{h, o}, s_{t+1}^{h, o}, ..., s_{t+n}^{h, o}$ in an auto-regressive manner. The history observations are concatenated with the PointNet feature to feed into the transformer, and the training is performed as supervised learning:
\begin{equation*}
\mathcal{L}(s_{t}^h, s_{t}^o) = \sum_{t=1}^{\Omega} \alpha\Vert s_{t}^h - \hat{s_{t}^h} \Vert^2 + \beta \Vert s_{t}^o - \hat{s_{t}^o} \Vert^2.
\end{equation*} We use $n=10$ in our experiments. The predicted observations $s_{t+n+1}^{h, o}$ are taken as input again for the next prediction until the length of the predicted sequence reach the sequence length $\Omega$ in datasets. We utilize a single, unified model for all datasets, adapting the input as necessary. For the bimanual ARCTIC dataset~\cite{fan2023arctic}, both hands are directly fed into the network. To accommodate data from Dex-YCB~\cite{chao2021dexycb}, we apply zero-padding to align input dimensions. In practice, we use $\alpha=1, \beta = 2$. $\beta$ is larger than $\alpha$ since the dimensions that belong to the object in the input have a lower ratio than the hand.

During inference on real robots, we regard robot hands as subcollections of human hands and directly apply MANO estimation on them. We extract 3D coordinates of robot joints $J_{Robot, i}$ (semantically corresponding to a subset $J_{MANO, i}$ of 21 joints ($N=21$) in MANO), then modify FrankMocap~\cite{rong2020frankmocap} by replacing the original optimization goal
\begin{equation*}
\sum_{i=1}^{N} || \Pi( (R, T) \cdot J_{MANO, i}(\theta, \beta) ) - J_{2D, i} ||^2 + L_{reg}(\theta, \beta)
\end{equation*}
with
\begin{equation*}
\mathcal{L}_{align} = \sum_{i=1}^{N} ||J_{MANO, i}(\theta, \beta) - J_{Robot, i} ||^2 + L_{reg}(\theta, \beta),
\end{equation*}
using robot proprioception and forward kinematics to get 3d joint positions $J_{Robot, i}$, instead of projecting ($\Pi( (R, T)$) the 3d joints $J_{MANO, i}$ to 2D and optimize in 2D. During optimization, shape parameters $\beta$ will be adaptively adjusted for different limb lengths, making it compatible with different types of robot hands with distinct sizes. During inference time, the object point cloud is attained by applying farthest point sampling (FPS)~\cite{qi2017pointnet++} to the object mesh, which is placed under 6DoF object pose estimation by model-based tracking in FoundationPose~\cite{wen2024foundationpose}.

\subsection{Robot-Object Trajectories Extraction}
The predicted hand motions $s_{t}^h$ are further converted to motions of a robotic hand $s_{t}^R$ using optimization-based hand motion retargeting. Following AnyTeleop~\cite{qin2023anyteleop}, the loss function minimizes the distance between the robot and human fingertip positions:
\begin{equation*}
    \mathcal{L}_{\text{retarget}}(q_t) = \sum_{i=1}^{5} \Vert x_t^i - f_i(q_t) \Vert^2 + \beta \Vert q_t - q_{t-1} \Vert^2,
\end{equation*}

subject to:
\begin{equation*}
    q_l \leq q_t \leq q_u,
\end{equation*}
where $q_t$ is the robot's joint configuration at time $t$, and $f_i(q_t)$ is the corresponding robot fingertip position calculated via forward kinematics to match the human's fingertip position, $x_t^i$. The robot's joints are constrained by lower ($q_l$) and upper ($q_u$) bounds. To ensure temporal smoothness, an additional penalty term is weighted by $\beta$. We solve this optimization problem using the NLopt solver \cite{qin2023anyteleop}. In practice, we set $\beta=1$, with $q_l=0^{\circ}$ and $q_u=90^{\circ}$.

\subsection{Trajectory-Guided Reinforcement Learning}
Due to the embodiment gap, the retargeted robot trajectories are imperfect and require further refinement. However, these predicted trajectories provide crucial guidance for learning robotic skills via reinforcement learning (RL). We represent the task using the Markov Decision Process (MDP) framework $\mathcal{M} = (\mathcal{S}, \mathcal{A}, P, r, \gamma)$, where the components correspond to the state space ($\mathcal{S}$), action space ($\mathcal{A}$), transition model ($P$), reward function ($r$), and discount factor ($\gamma$), respectively.

We train a state-based skill policy, $\pi_i$. The policy's input state $s_t \in \mathcal{S}$ is composed of robot joint positions and 6D object poses. The output action $a_t \in \mathcal{A}$ consists of delta end-effector poses and delta hand joint positions.
\begin{equation*}
    a_t \sim \pi(a_t \mid s_t).
\end{equation*}
At each time step $t$ of the interaction, the environment transitions to $s_{t+1}$ based on $P(s_{t+1} \mid s_t, a_t)$ and provides a reward $r_t$. During control, the target end-effector pose is converted to arm joint positions via inverse kinematics (IK) using Pinocchio \cite{carpentier2019pinocchio}. Each policy $\pi_i$ from a set $\Pi$ is optimized to maximize the expected cumulative reward, conditioned on a reference trajectory $\tau_i$ from a human demonstration with task horizon $H$:
\begin{equation*}
    \pi^* = \arg\max_{\pi} \mathbb{E} \left[ \sum_{t=0}^{H} \gamma^t r(s_t, a_t \mid \tau_i) \right].
\end{equation*}

Our reward function consists of four items:

\begin{equation*}
    R=\lambda_1 R_{f} + \lambda_2 R_{o} +\lambda_3 R_{\text{contact}}  + \lambda_4 R_{\text{success}}.\label{eq:r1}
\end{equation*}The first term, robot following reward $R_f=\beta_1 R_{j} + \beta_2 R_{\text{ee}, x} + \beta_3 R_{\text{ee}, r} \label{eq:r2}$, encourages the robot to follow reference hand motions.

The first component $R_j=1 - \tanh{(\alpha_1 \Vert q^t - \hat{q}^{t^\prime} \Vert)}$ encourage the robot hand to follow reference joint positions, where $q_t$ and $\hat{q}_{t^\prime}$ are robot hand joint positions at timestep $t$ and reference robot hand joint positions at corresponding timestep $t^\prime$. The second one $R_{\text{ee}, x}=1-\tanh{(\alpha_2 \Vert x_{\text{ee}}^t - \hat{x}_{\text{ee}}^{t^\prime}\Vert)}$ and the third one $R_{\text{ee}, r}=1-\tanh{(\alpha_3 \phi (\theta^t, \hat{\theta}^{t^\prime}))}$ constrains robot End-Effector (EE) pose where $x_{\text{ee}}$ are robot EE position and $\phi$ computes the angular distance between robot EE orientation $\theta^t$ with reference $\hat{\theta}^{t^\prime}$.

The second term is an \textbf{object following reward}. Instead of using human-designed task-specific reward like lifting reward in previous work~\cite{chen2024vividex, shi2025learning}, we utilize the predicted object states $S_{t}^o$ as guiding signals to encourage the robots to manipulate the object in the way that is in the dataset, which is more generalizable to different tasks:
\begin{equation*}
    R_{o} = e^{-\alpha_{4}\Vert \mathbf{T_{t}^o-\hat{\mathbf{T_{t}^o}}} \Vert_{2}}+e^{-\alpha_{5}\Vert {log(\hat{\mathbf{q}}_{t}^{o}}^{-1}\otimes \mathbf{q}_{t}^{o}) \Vert_{2}},
\end{equation*}
where $\log(\cdot)$ is the Logarithmic Map which converts a quaternion to its axis-angle representation, and $\otimes$ denotes the Hamilton product.

We encourage finger exploration through $R_{\text{contact}}$
, which gives a reward only when the thumb finger and at least one other finger contact with the object. The contact information is derived from the collision check in SAPIEN simulation~\cite{xiang2020sapien}. In obstacle avoidance tasks, we further add a large penalty when any part of the hand or the arm contact the obstacle, forcing the policy to explore a plausible route to avoid obstacles.

We give a large reward $R_{\text{success}}$ when task is finished with success.
We use PPO~\cite{schulman2017proximal} for state-based policy learning with delta end-effector pose control, where we randomize initial object position within $\pm 5$ cm in $xy$-plane and rotation within $\pm 30^\circ$ around the $z$-axis.

We carefully tuned the reward weights. In our experiments, we use
$\lambda_1=\lambda_2=\lambda_3=1,\lambda_4 =15,\lambda_5=3, \beta_1=0.1, \beta_2=\beta_3=1,\alpha_1=\alpha_2=\alpha_3=\alpha_4=\alpha_5=\alpha_6=1$ where $R_{success}=1$ upon task success and $R_{success}=0$ otherwise. In \textit{YCB grasping with obstacle} task, the weights of the contact reward are modified to be -20 when there is any contact between the robot hand links or the robot arm links with the obstacle.

\subsection{Sim-to-Real Transfer}
We have several designs to overcome the control gap and perception gap between the simulation and the real world, enabling zero-shot Sim-to-Real transfer. Before training the reinforcement learning policy, we perform system identification~\cite{ma2025learning} to mitigate the control gap. Specifically, we control the simulated robot and the real robot to move in the same direction in the same time interval and minimize the error between the two by adjusting the parameters of the simulated robot, including PID parameters and force limits for each joint motors. For different robot hand embodiments, we tune the control parameters for each to minimize the control gap when switched to another embodiment. Besides, we adopt domain randomization~\cite{tobin2017domain} by artificially adding random noise to the observations as well as adding simulation parameter randomizations to friction, object scale, and object weights in the simulation to improve the robustness of the trained policies to the real-world noise.

\begin{figure*}[t]
\centering
\includegraphics[width=\textwidth]{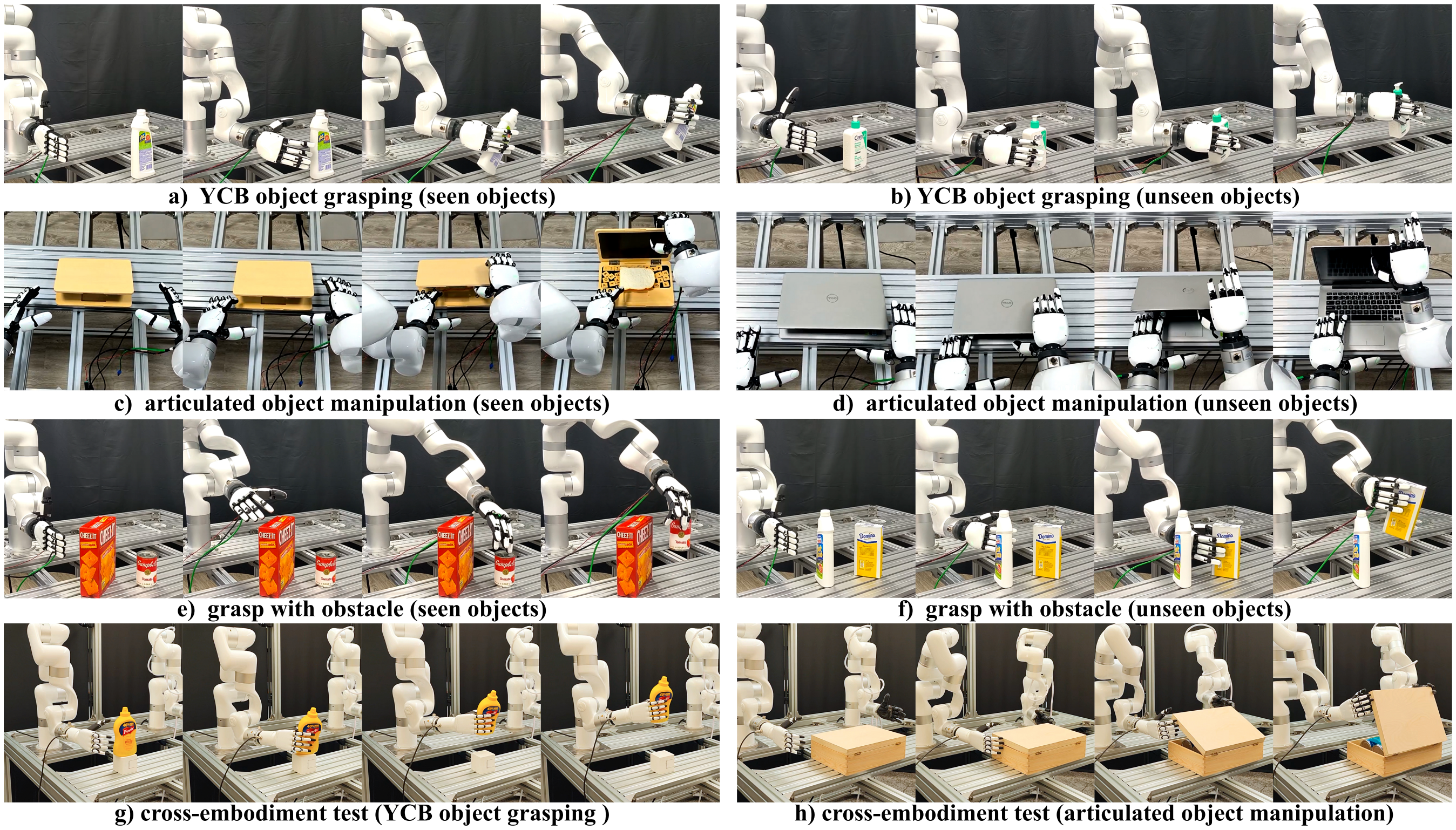}
\caption{Qualitative results from real-world experiments.}
\label{fig: experiment results overview}
\end{figure*}

\section{Experiments}
\subsection{Experiment Setting}
As shown in Fig.~\ref{fig: experiment results overview}, our experiments cover the following 4 settings:
\begin{itemize}
    \item 1) \textit{YCB object grasping}: the success is defined when the robot hand successfully grasps a target YCB object placed on the platform and lifts it up to 10 cm above its initial height.
    \item 2) \textit{Articulated object manipulation}: the task success is defined as the hinged object being rotated more than 30 degrees along the axis.
    \item 3) \textit{Grasping with obstacles}: the distinction between this task and a standard grasping task is the deliberate introduction of an obstacle along the robot's path towards the target object. An additional constraint is that any part of the robotic arm and dexterous hand must not collide with the obstacle at any point during the entire process of grasping the target object.
    \item 4) \textit{Cross-embodiment test}: we replace the right-side robot hand with a different model. We then reuse the motion predictor to generate human-object interaction trajectories and apply retargeting adapted to the specific robot configuration. Finally, we perform trajectory-guided reinforcement learning to learn the skills, using identical reward terms and parameters for different robot hands.
\end{itemize}

\subsection{Abbreviation:} In the following paragraphs, the terms \textbf{seen} and \textbf{unseen} (as applied to objects or settings) refer to whether that configuration was used to train our HOI motion predictor and the corresponding motion extractors for the baseline methods.
\subsection{Simulation Configuration}For simulation, we employ SAPIEN3 \cite{xiang2020sapien} and OpenAI-Gymnasium \cite{towers2024gymnasium} to enable GPU-parallel training and rendering, we have 1024 environments running in parallel on single NVIDIA RTX 4090 GPU (24GB) and i7-14700K CPU, with average steps per second at 10000, and training for 1M steps for the policies to converge. Parameters used in training the PPO reinforcement learning policies are shown in Tab.~\ref{tab:ppo}.

\begin{table}[t]
\centering
\fontsize{9pt}{10.8pt}\selectfont
\renewcommand{\arraystretch}{1.12}
\begin{tabular*}{\columnwidth}{@{\extracolsep{\fill}}lc@{}}
\toprule
Parameter & Value \\
\midrule
Maximum steps & $10^{6}$ \\
Bootstrap at done & True \\
Discount factor ($\gamma$) & 0.95 \\
GAE parameter ($\lambda$) & 0.9 \\
Learning rate & $3\times10^{-4}$ \\
Parallel environments & 1024 \\
\bottomrule
\end{tabular*}
\caption{Training parameters of PPO policies.}
\label{tab:ppo}
\end{table}
\subsection{Real-World Robot Setup} Our bimanual robot platform consists of two xArm-7 arms, each mounted with a PSYONIC Ability hand, a ROBOTERA XHand1 or an Inspire robot hand. Object pose estimation is performed using model-based tracking with FoundationPose~\cite{wen2024foundationpose} based on RGB-D data from a RealSense D435i camera, which was calibrated using EasyHec++~\cite{hong2024easyhec++}.
\subsection{Baselines}
For YCB object grasping tasks, we compare our results with existing SOTA methods. They are ViViDex~\cite{chen2024vividex}, AdaDexGrasp~\cite{shi2025learning}, ManipTrans~\cite{li2025maniptrans}, HOP~\cite{singh2024hand} using RL for finetuning, and PPO w/o following, which substitutes following reward $R_{f}$ and $R_{o}$ using a reaching reward:
\begin{equation*}
    R_{reaching}=\lambda_{5}(1 - \tanh{(\alpha_6 \Vert T_{t}^{TCP} - \hat{T}^{o} \Vert))},
\end{equation*}
 where $T_{t}^{TCP}$ is the translation of the robot TCP pose, which refers to the pose of the palm of the robot hand, and $\hat{T}^{o}$ is the global translation of the object. When there is a novel object configuration, ViViDex~\cite{chen2024vividex} and AdaDexGrasp~\cite{shi2025learning} use existing reference trajectories in the dataset. For ARCTIC articulated object manipulation, we compare with Obj-Dex~\cite{chen2024object}, ManipTrans~\cite{li2025maniptrans}, and PPO w/o following. For grasp with obstacle tasks, we also compare ours with HOP~\cite{singh2024hand}. We use the PSYONIC Ability hand, ROBOTERA XHand1 or Inspire robot hand for the cross-embodiment test and compare the results. All the simulation experiments are tested with 5 seeds, and we report the average success rates in policy evaluation using 100 trials each after the convergence of RL training. All the real-world robot experiments are repeated 20 times to get success rates.

\subsection{YCB Object Grasping}
We test the methods on different YCB objects (5 objects in Dex-YCB datasets~\cite{chao2021dexycb}, 5 objects from YCB datasets~\cite{calli2015benchmarking} which are not included in Dex-YCB datasets) and 5 different initial object poses with randomized position and rotation inside the size of an A4 paper. The task success is defined as the object being grasped and lifted up over 10 cm. Success rates in simulation and real-world robots are reported in Tab.~\ref{tab:grasp-sim}. ViViDex and AdaDexGrasp struggle to handle non-DexYCB objects. ManipTrans has difficulty dealing with novel initial object poses. Our method demonstrates consistent performance across different task configurations.

\begin{table*}[t]
\centering
\fontsize{9pt}{10.8pt}\selectfont
\renewcommand{\arraystretch}{1.15}
\setlength{\tabcolsep}{3pt}
\begin{tabular*}{\textwidth}{@{\extracolsep{\fill}}lcccccccc@{}}
\toprule
Method & \multicolumn{4}{c}{Simulation} & \multicolumn{4}{c}{Real world} \\
\cmidrule(lr){2-5}\cmidrule(lr){6-9}
 & DexYCB & Non-DexYCB & Novel poses & Avg. & DexYCB & Non-DexYCB & Novel poses & Avg. \\
\midrule
ViViDex~\cite{chen2024vividex} & 0.87 & 0.37 & \textbf{0.75} & 0.66 & 0.35 & 0.40 & 0.25 & 0.33 \\
AdaDexGrasp~\cite{shi2025learning} & 0.82 & 0.21 & 0.64 & 0.56 & 0.80 & 0.20 & \textbf{0.65} & 0.55 \\
ManipTrans~\cite{li2025maniptrans} & 0.61 & 0.28 & 0.03 & 0.31 & 0.60 & 0.30 & 0.05 & 0.32 \\
HOP~\cite{singh2024hand} & 0.63 & 0.59 & 0.54 & 0.59 & 0.60 & 0.45 & 0.40 & 0.48 \\
PPO w/o following~\cite{schulman2017proximal} & 0.37 & 0.42 & 0.27 & 0.35 & 0.30 & 0.40 & 0.20 & 0.30 \\
\midrule
\textbf{Ours} & \textbf{0.90} & \textbf{0.86} & \textbf{0.75} & \textbf{0.84} & \textbf{0.85} & \textbf{0.80} & \textbf{0.65} & \textbf{0.77} \\
\bottomrule
\end{tabular*}
\caption{Success rates for \textit{YCB object grasping} in simulation and on real robots. The novel-pose condition uses DexYCB objects.}
\label{tab:grasp-sim}
\label{tab:grasp-real}
\end{table*}

\begin{table*}[t]
\centering
\fontsize{9pt}{10.8pt}\selectfont
\renewcommand{\arraystretch}{1.15}
\setlength{\tabcolsep}{3pt}
\begin{tabular*}{\textwidth}{@{\extracolsep{\fill}}lcccccc@{}}
\toprule
Method & \multicolumn{3}{c}{Simulation} & \multicolumn{3}{c}{Real world} \\
\cmidrule(lr){2-4}\cmidrule(lr){5-7}
 & ARCTIC & Daily & Avg. & ARCTIC & Daily & Avg. \\
\midrule
ManipTrans~\cite{li2025maniptrans} & 0.61 & 0.00 & 0.31 & 0.55 & 0.00 & 0.28 \\
Obj-Dex~\cite{chen2024object} & 0.74 & 0.05 & 0.40 & \textbf{0.70} & 0.00 & 0.35 \\
PPO w/o following~\cite{schulman2017proximal} & 0.00 & 0.00 & 0.00 & 0.00 & 0.00 & 0.00 \\
\midrule
\textbf{Ours} & \textbf{0.79} & \textbf{0.53} & \textbf{0.66} & \textbf{0.70} & \textbf{0.35} & \textbf{0.53} \\
\bottomrule
\end{tabular*}
\caption{Success rates for \textit{articulated object manipulation} in simulation and on real robots. ARCTIC and Daily denote objects from the ARCTIC dataset and daily-life objects, respectively.}
\label{tab:hinge-sim}
\label{tab:hinge-real}
\end{table*}

\begin{table*}[t]
\centering
\fontsize{9pt}{10.8pt}\selectfont
\renewcommand{\arraystretch}{1.15}
\setlength{\tabcolsep}{3pt}
\begin{tabular*}{\textwidth}{@{\extracolsep{\fill}}lcccccc@{}}
\toprule
Method & \multicolumn{3}{c}{Simulation} & \multicolumn{3}{c}{Real world} \\
\cmidrule(lr){2-4}\cmidrule(lr){5-7}
 & \shortstack{DexYCB\\obstacle} & \shortstack{3D-printed\\obstacle} & Avg. & \shortstack{DexYCB\\obstacle} & \shortstack{3D-printed\\obstacle} & Avg. \\
\midrule
ViViDex~\cite{chen2024vividex} & 0.17 & 0.38 & 0.28 & 0.15 & 0.40 & 0.28 \\
AdaDexGrasp~\cite{shi2025learning} & 0.55 & 0.67 & 0.61 & \textbf{0.55} & 0.60 & 0.58 \\
ManipTrans~\cite{li2025maniptrans} & 0.00 & 0.00 & 0.00 & 0.00 & 0.00 & 0.00 \\
HOP~\cite{singh2024hand} & 0.32 & 0.41 & 0.37 & 0.25 & 0.40 & 0.33 \\
PPO w/o following~\cite{schulman2017proximal} & \textbf{0.58} & 0.63 & 0.61 & 0.50 & 0.55 & 0.53 \\
\midrule
\textbf{Ours} & 0.57 & \textbf{0.74} & \textbf{0.66} & \textbf{0.55} & \textbf{0.70} & \textbf{0.63} \\
\bottomrule
\end{tabular*}
\caption{Success rates for \textit{grasping with obstacles} in simulation and on real robots. Real-world collisions are detected using xArm's built-in collision detector.}
\label{tab:obstacle-sim}
\label{tab:obstacle-real}
\end{table*}

\subsection{Articulated Object Manipulation}
We test the performance of the methods on 3 objects in ARCTIC~\cite{fan2023arctic} dataset, which are \textit{wooden laptop}, \textit{wooden box}, \textit{mixer}, and 3 objects in our daily life: \textit{real laptop}, \textit{larger box}, \textit{toy coffee machine}. The task success is defined as the hinged object rotating more than 30 degrees along the axis. We report success rates in simulation and real-world robots in Tab.~\ref{tab:hinge-sim}, where ~\cite{li2025maniptrans},~\cite{chen2024object} fail to handle the novel daily objects, while we still observe success in ours in simulation and real robots.

\subsection{Grasping with Obstacle}
The distinction between this task and a standard grasping task is the deliberate introduction of an obstacle along the robot's path towards the target object. While the obstacle's mesh is known, its type and position are variable. An additional constraint is that the robotic arm and dexterous hand must not collide with the obstacle at any point during the process of grasping the target object; otherwise, the task is considered a failure. We select 5 objects in the DexYCB datasets and 5 3D-printed object to make up our obstacle library and choose one obstacle from it each time. In this experiment, a large penalty item $R_{obstacle}$ is introduced to each method when any part of the hand or the arm contacts the obstacle during RL training, forcing the policy to explore a plausible route to avoid obstacles. Besides, we selectively filter out all time frames where the dexterous hand meshes from reference trajectories collide with obstacles using cuRobo~\cite{sundaralingam2023curobo} motion planning, disabling the following reward in these frames for ours and baselines. We report success rates in simulation and real-world robots in Tab.~\ref{tab:obstacle-sim}.

\begin{figure}[t]
\centering
\includegraphics[width=0.47\textwidth]{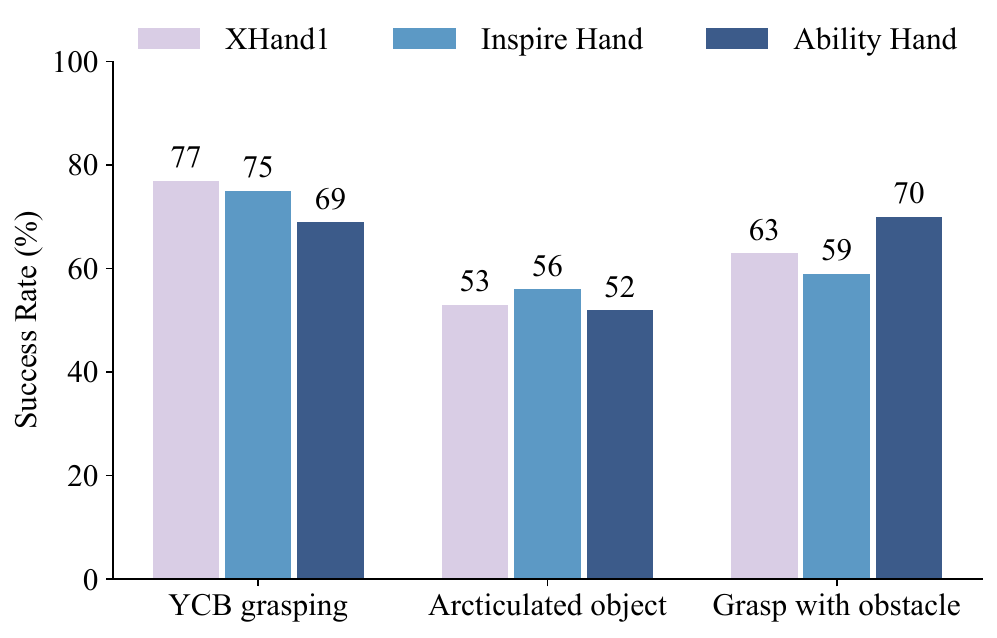}
\caption{Cross-embodiment test on 3 robot hands.}
\label{fig:validate}
\end{figure}

\subsection{Cross-Embodiment Test}
To validate the generalizability of our method across different robotic platforms, we replace the right-side robot hand in the robot setting with another robot hand. We then re-execute the retargeting and reinforcement learning algorithms based on the trajectories generated by our motion predictor and evaluated the performance on various tasks in the real-world robots. See Fig.~\ref{fig:validate}, our methods perform well across different embodiments of dexterous hands, showing consistent performance in different tasks.

\begin{figure}[t]
\centering
\includegraphics[width=0.47\textwidth]{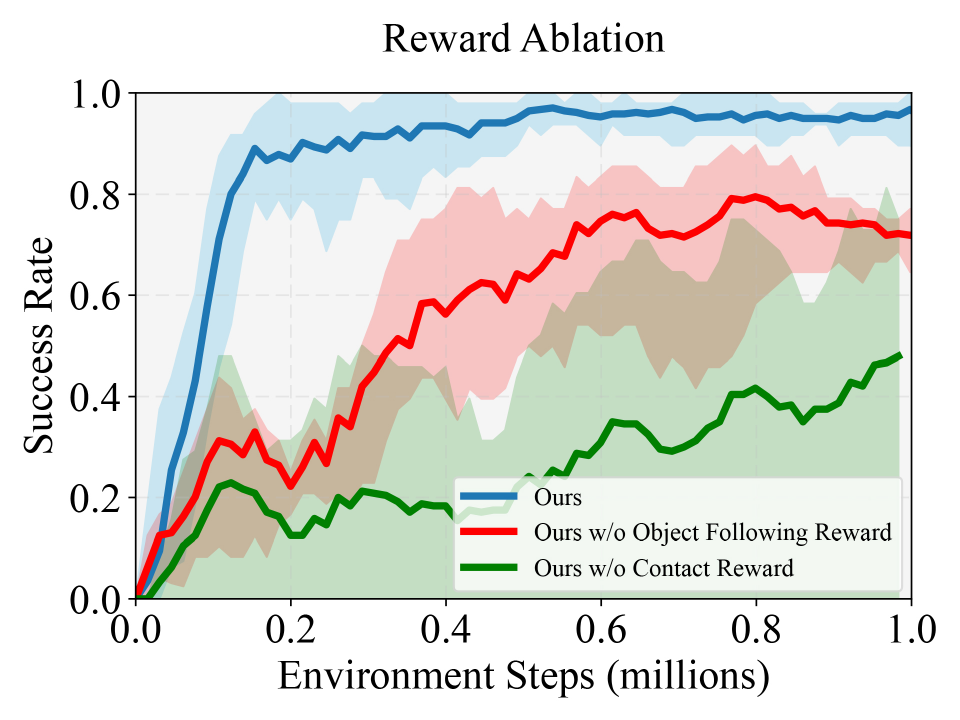}
\caption{Learning curves for reward ablation (Solid lines: the mean over all seeds. Shaded areas: standard deviation).}
\label{fig:curve}
\end{figure}

\subsection{Ablation Study on the Reward Design}
To validate the effectiveness of our reward design, we conduct ablation studies by: 1) removing the object following term $R_{o}$, and 2) removing the contact reward $R_{contact}$. We evaluate these on the YCB object grasping task using 5 objects (5 seeds each) from the DexYCB dataset and report the average success rates in Fig.~\ref{fig:curve}. We observe that when $R_{o}$ is removed, sample efficiency decreases dramatically. When $R_{contact}$ is removed, training becomes unstable, resulting in lower success rates. These results demonstrate that the object following reward $R_{o}$ is critical for ensuring stable and efficient policy learning.

\section{Conclusion}
We propose a hierarchical framework for learning dexterous manipulation from object-centric human motion priors to achieve generalizable manipulation, enabling zero-shot Sim-to-Real transfer. Our method outperforms existing baselines in YCB object grasping, articulated object manipulation, and grasping with obstacles in terms of success rates and generalizations to unseen objects. We demonstrate the importance of object following reward through ablation studies. We also validate the cross-embodiment potentials of our methods.

\bibliography{aaai2026}

\end{document}